# Machine Translation Approaches and Survey for Indian Languages

Nadeem Jadoon Khan[*]    Waqas Anwar*    Nadir Durrani[**]

*COMSAT University

**University of Edinburgh

**ABSTRACT:** In this study, we present an analysis regarding the performance of the state-of-art Phrase-based Statistical Machine Translation (SMT) on multiple Indian languages. We report baseline systems on several language pairs. The motivation of this study is to promote the development of SMT and linguistic resources for these language pairs, as the current state-of-the-art is quite bleak due to sparse data resources. The success of an SMT system is contingent on the availability of a large parallel corpus. Such data is necessary to reliably estimate translation probabilities. We report the performance of baseline systems translating from Indian languages (Bengali, Guajarati, Hindi, Malayalam, Punjabi, Tamil, Telugu and Urdu) into English with average 10% accurate results for all the language pairs.

## 1. Introduction

In this section, a brief background of Machine Translation is given. An overview of Machine Translation (MT) approaches is also discussed with the SMT approach being used to carry out this work. Indian languages that we selected for this work are also discussed briefly.

### 1.1. Machine Translation

Machine Translation (MT) can be defined as an automated system that analyses text from a Source Language (SL), applies some computation on that input and produces equivalent text in a required target language (TL) ideally without any kind of human intervention.
It is one of the most interesting and the hardest problem in the field of NLP (Koehn, 2010). The two challenges in machine translation are adequacy and fluency. The former is to develop a system that adequately represents the ideas expressed in the source language into the target language. The latter is to represent those ideas grammatically. The common approaches to machine translation are the rule-based approach and corpus-based approach.
In the rule-based approach, the text in the source language is analyzed using various tools such as: a morphological parser and analyser and then transformed into an intermediate representation. A set of rules are used to generate the text in target language of this intermediate representation. A large number of rules are necessary to capture the phenomena of natural language. These rules transfer the grammatical structure of the source language into target language. As the number of rules increases, the system become more complicated (Islam et al., 2010) and slow to translate. Formulation of a large number of rules is a tedious process and requires years of effort and linguistic analysis.
In another approach, large parallel and monolingual corpora are used as source of knowledge. This approach can be further divided into statistical approach and example-based approach. In the statistical approaches, target text is generated and scored through a statistical model, the parameters of which are learned from parallel corpus. Here, MT is also seen as a decision problem, a better target language phrase id is decided from the given source language. Bayes rule and statistical decision theory are used to solve this decision problem. Statistical decision theory and Bayesian decision rules are used to minimize errors of decision. SMT (Koehn, 2010) gives better results if additional training data are available.

Figure 1: Architecture of a typical SMT system

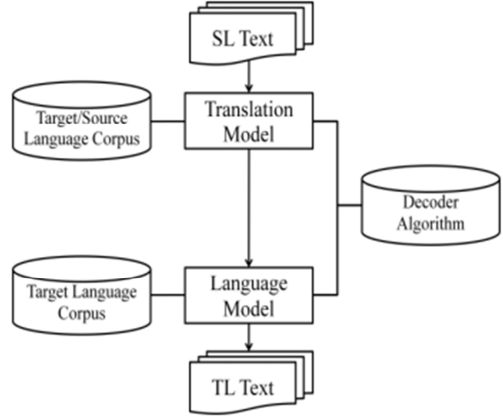

SMT is superior to rule-based and example-based systems in that it does not require human interpenetration and can build a translation system in an unsupervised manner directly from the training data. With the rapid proliferation of internet and increasing availability of data, SMT is currently the most popular and prevalent paradigm. SMT can be represented by different models and a basic architecture of simple SMT system model is shown in Figure 1. An arrow from Translation model to Language model shows that the Language model contains the target side corpora as well. The arrow from Language model to Translation text shows that the fluency of the translation depends upon the quality of Language Model. In this study we use Phrase-based SMT model and an overview of this model is given in next section.

### 1.1.1. Phrase-based Model

In our experiment, we have used the Phrase-based SMT Models (Koehn et al., 2003; Och & Ney, 2004) and evaluated their performance on the morphologically rich Indian languages. Phrase-based models are used to translate phrases of one or more words as atomic units (Koehn, 2010). These models divide the input sentence into phrases, produce the target phrases and at the end reordering of these phrases is done. Phrase-based models memorize local dependencies such as short reordering, idiomatic collocations, insertions and deletions.

Phrase-based models are based on the noisy channel model introduced by (Shannon, 1948) in the information theory. Given a source sentence *F*, the objective is to find a target sentence *E*, which maximizes the likelihood of two components, the translation (or adequacy) and the language (or fluency model).

Every sentence *F* is an arrangement of words symbolized as $f_1^J = f_1...f_j....f_J$ is decoded into a sentence E of target language, and symbolized as $e_1^I = e_1...e_i...e_I$. The objective is to find a target sentence that maximizes the model:

$$ê_1^I = \text{argmax } P(e_1^I | f_1^J) \quad (1)$$

For decoding sentence $f_1^J$ into sentence $e_1^I$, we require to calculate $P(e_1^I | f_1^J)$, the translation model probability. Using Bayes theorem we can decompose the above equation as:

$$P(e_1^I | f_1^J) = \frac{P(f_1^J | e_1^I) \cdot P(e_1^I)}{P(f_1^J)} \qquad (2)$$

Subsequently our goal is to get the most out of general probable translation hypotheses for the specified source sentence $f_1^J$. Equation 2 will be computed for every sentence in Language *E*. But $P(f_1^J)$ is not modified for every translation hypothesis. Therefore we can neglect the denominator $P(f_1^J)$ from the Equation 2.

$$\hat{e}_1^I = argmax\, P(f_1^J | e_1^I) \cdot P(e_1^I) \qquad (3)$$

The model of the likelihood distributed for the first term in Equation 3 ($P(f_1^J | e_1^I)$), probability of translation (*f,e*)) is called **Translation Model**, and the distribution of $P(e_1^I)$ is called the **Language Model**.

## 1.2. Language Selection

We selected 8 commonly spoken languages in the sub-continent, parallel corpus of which was available to test.

**Bengali (Bangla) :** Bengali is the national language of Bangladesh and one of the officially spoken languages of India. More than 21 million people speak Bengali either as their first or second language (Britannica, 2014). There are roughly 10 million native speakers of Bengali in Bangladesh and around 85 million in India in the states like west Bengal, Assam and Tripura. Bengali is also known as Bangla and it is associated with Indo-Iranian family. Like most languages it is also written from left to right. Its sentence structure is similar to English i.e. SOV (Subject Object Verb). All letters are written in same case and there are no capital letters. The source of its punctuation is English language of 19th century.

**Gujarati:** It is a member of Indo-Aryan branch of languages. 46 million people in the Indian state of Gujarat speak Guajarati (Britannica, 2014). Evolution of Gujarati language took place in 12th century. Gujarati declension is considerably complicated. It contains 3 genders masculine, feminine, and neuter and 2 numbers singular and plural. For nouns it has three cases nominative, oblique and agentive locative. It is written from left to right with writing style SOV.

**Hindi**: It is the national and official language of India. 425 million people speak Hindi as their first language while more than 12 million people as their second (Britannica, 2014). Outside India, some communities in South Africa, Mauritius, Bangladesh, Yemen, and Uganda also communicate in Hindi language. Hindi is a-member of the-Indo-Aryan-group within the Indo-Iranian-branch of the-Indo-European-language family. Like in Persian, Hindi adjectives do not change as a result of number change in noun. Its preposition is similar to English. Unlike others Sanskrit based languages like Guajarati it has only two genders i.e. masculine and feminine. Case marking in Hindi is simple due to Persian influence and reduces it to direct form and an oblique form. Case relations are shown postpositions. Like many languages it is also written from left to right but its writing style is SOV. Modern standard Hindi evolved from the interaction of Muslim from Afghanistan, Iran, Turkey, Central Asia, and elsewhere.
Due to Persian influence Hindi borrowed some part of vocabulary from Persian language such as dresses (e.g. پاجامہ, pajama, (Trouser); چادر ،chadar (Sheet),  cuisine (e.g. قورمہ,korma; کباب,kebab), cosmetics

(e.g. صابن, sabun (soap); حنا,hina, hen-na), furniture (e.g., كرسى, kursi (chair); ميز,maiz (table)), construction (e.g., ديوار (wall)).

A large number of adjectives and their nominal derivatives(e.g.,-abad-inhabitedand-abadi-population), and a wide range of other items and concepts are so much a part of the Hindi language that purists of the post-independence period have been unsuccessful in purging them. While borrowing Persian and Arabic words, Hindi also borrowed-phonemes, such as /f/ and /z/, though these were sometimes replaced by /ph/ and /j/. For instance, Hindi renders the word for force as either-zor-or-jor-and the word for sight as-nazar-or-najar. In most cases the sounds /g/ and /x/ were replaced by /k/ and /kh/, respectively. Contact with the-English language-has also enriched Hindi. Many English words, such as-button,-pencil,-petrol, and college-are fully assimilated in the Hindi lexicon.

**Malayalam:** Malayalam is also a widely spoken language in India, mainly in the state of Kerala where it is an official language. In Tamil Nadu and Karnataka, few societies communicate in Malayalam language. It belongs to South Dravidian which is sub part of Dravidian language. Around 35 million people speak this language (Britannica, 2014).There exist different slangs between social caste lines which causes diglossia i.e. difference between formal, literary and colloquial forms of speech. Like other Dravidian languages it also has a series of retroflex constants **(/ḍ/, /ṇ/, and /ṭ/)** pronounce by touching the tip of tongue to the roof of the mouth. Its writing style is SOV and has nominative accusative case marking pattern. It has three genders i.e. masculine, feminine and neuter. Inflection is generally marked via suffixation. Unlike other Dravidian languages, Malayalam inflects its finite verb only for tense—not for person, number, or gender.

**Punjabi (Panjabi):** It is a member of the-Indo-Aryan-subdivision of the-Indo-European language-family. More than 10 million people speak this language (Britannica, 2014) in the domain that was discordant between Pakistan and India during cleave. This language is officially added in Indian constitution. Some small societies in UAE, UK, USA, Canada, South Africa and Malaysia speak Punjabi. It is of two miscellanies; one is western which is known as Lahnda and second is eastern known as Gurmukhi. There are two ways to write Punjabi, one is by Perso-Arabic script and other is by Gurmukhi alphabets which were conceived by Sikh Guru Angad (1539-52) rules for scriptural use. Its writing style is SOV and written from left to right (Gurmukhi) and right to left (Perso-Arabic).

**Tamil:** Tamil is the member of Dravidian language and is the official language of the Tamil Nadu state. It is also the official language in Sri Lanka and Singapore and is also spoken by many people is Malaysia, Mauritius, Fiji and South Africa. In 2004, it was declared as classical language of India which means it met three criteria its origins are ancient; it has an independent tradition; and it possesses a considerable body of ancient literature. Around 66 million people speak Tamil language (Britannica, 2014).
Three times, changes occurred in grammatical and lexical form of this language, Old Tamil (from about 450 BCE to 700 CE), Middle Tamil (700 CE 1600 CE), and Modern Tamil (from 160 CE onwards). Its writing system developed from Brahmi script. Over the time its letters changed shapes until 16th century CE when printing was introduced and its shape stabilized. The major addition to the alphabet was the incorporation of Grantha-letters to write unassimilated Sanskrit words, although a few letters with irregular shapes were standardized during the modern period. . A script known as Vatteluttu (Round Script) is also in common use. With time, changes in the way of speaking this language occurred. Tamil language spoken in India is different from that which is spoken in Sri Lanka. Variation takes place in different regions. Within-Tamil Nadu-there are phonological differences between the northern, western, and southern speech.

**Telugu**: Telugu is one of the most spoken languages among the Dravidian language family. In south eastern part of India, people communicate in this language. In Andhra Pradesh it is the official language. Worldwide, 75 million people speak Telugu language (Britannica, 2014). The oldest material belonging to this language is of 575 CE. The Telugu script is used for writing Telugu, which is derived from Calukya Dynasty. Telugu is written from left to right. Visually, it differs from many of the North Indian scripts in that the letters have a rounded base.

**Urdu** is also a member of the Indo-Aryan group within the Indo-European family of languages. Urdu is the national language of Pakistan while it is officially recognized language in Indian constitution as well. More than 100 million people (Britannica, 2014) within Pakistan and India speak in Urdu. Apart from these two nations Urdu is also spoken by the immigrants and in small societies in UK, USA and UAE. Urdu and Hindi are bilaterally audible. This language developed and stemmed from Indian subcontinent therefore it is similar to Hindi. Due to similarity in phonics and grammar they seem like one language but there sources are different. Urdu is lent from Arabic and Persian while Hindi is borrowed from Sanskrit that is why they are treated as maverick languages. There is a huge difference in their writing style. Urdu script is an altered and revised form of Perso-Arabic scripts while Hindi script is a modified form of Devanagari script. Urdu and Hindi sound similar except few variations in short vowel allophones. Urdu withholds a full set of aspirated stops. It is the property of both Indo Aryan as well as retroflex stops. Urdu does not retain the complete range of Perso-Arabic consonants, despite its heavy borrowing from that tradition. The largest number of sounds retained is among the spirants; a group of sounds uttered with a friction of breath against some part of the oral passage, in this case /f/, /z/, /zh/, /x/, and /g/. One sound in the stops category, the glottal /q/, has also been retained from Perso-Arabic. Grammatically Hindi and Urdu are same. Major difference between these two is Urdu is written from right to left while Hindi is written from left to right. Style of Urdu writing is SOV and exhibit split ergative behaviour. In Urdu, Perso Arabic prefixes and suffixes are more than Hindi. examples include the prefixes *dar-* 'in,' *ba-/baa-* 'with,' *be-/bila-/la-* 'without,' and *bad-* 'ill, miss' and the suffixes *-dar* 'holder,' *-saz* 'maker' (as in *zinsaz* 'harness maker'), *-khor* 'eater' (as in *muftkhor* 'free eater'), and *-posh* 'cover' (as in *mez posh* 'table cover').

## 1.3. Related Work

Initial research has been done to translate Indian languages, mostly focusing Hindi and Bengali. However, most of the focus is still rule-based because of the unavailability of parallel data to build SMT systems for these languages.
(Dasgupta et al., 2004) proposed an approach for English to Bangla MT that uses syntactic transfer of English sentences to Bangla with optimal time complexity. In generation stage of the phrases they used a dictionary to identify subject, object and also other entities like person, number and generate target sentences. (Naskar et al., 2006) presented an example-based machine translation system for English to Bangla. Their work identifies the phrases in the input through a shallow analysis, retrieves the target phrases using the example-based approach and finally combines the target phrases using some heuristics based on the phrase reordering rules from Bangla. They also discussed some syntactic issues between English and Bangla. (Anwar et al., 2009) proposed a method to analyze syntactically Bangla sentence using context sensitive grammar rules which accepts almost all types of Bangla sentences including simple, complex and compound sentences and then interpret input Bangla sentence to English using a NLP conversion unit. The grammar rules employed in the system allow parsing five categories of sentences according to Bangla intonation. The system is based on analyzing an input sentence and converting into a structural representation (SR). Once an SR is created for a particular sentence it is then converted to corresponding English sentence by NLP conversion unit. For conversion, the NLP conversion utilizes the corpus. (Islam et al., 2010) proposed a phrase-based Statistical Machine Translation (SMT) system that translates English sentences to Bengali. They added a transliteration module to handle OOV words. A preposition handling module is also incorporated to deal with systematic grammatical differences between English and Bangla. To measure the performance of their system, they used BLEU, NIST and TER scores. (Durrani et al., 2010) also made use of transliteration to aid translation between Hindi and Urdu which are closely related languages. (Roy &

Popowich, 2009) applied three reordering techniques namely lexicalized, manual and automatic reordering to the source and language in a Bangla English SMT system. (Singh et al., 2012) presented a Phrase based model approach to English-Hindi translation. In their work they discussed the simple implementation of default phrase-based model for SMT for English to Hindi and also give an overview of different Machine translation applications that are in use nowadays.

(Sharma et.al. 2011) presented English to Hindi SMT system using phrase-based model approach. They used human evaluation metrics as their evaluation measures. These evaluations cost higher than the already available automatic evaluation metrics. (Yamada & Knight, 2001) used methods based on tree to string mappings where source language sentences are first parsed and later operations on each node. (Eisner, 2003) presented issues of working with isomorphic trees and presented a new approach of non-isomorphic tree-to-tree mapping translation model using synchronous tree substitution grammar (STSG). (Li et al., 2005) first gave idea of using maximum entropy model based on source language parse trees to get n-best syntactic reordering's of each sentence which was further extended to use of lattices.

(Bisazza & Federico 2010) further explored lattice-based reordering techniques for Arabic-English; they used shallow syntax chunking of the source language to move clause-initial verbs up to the maximum of 6 chunks where each verb's placement is encoded as separate path in lattice and each path is associated with a feature weight used by the decoder.

(Jawad et al., 2010) presented complete study work for English to Urdu MT that uses factored based MT. In their work they discussed the complete divergence between two languages. Vocabulary difference between Urdu and English has been discussed. In their work they showed importance of factored based models when we got information about the morphology of both source and targeted language.

(Khan, et al., 2013) presented baseline SMT system for English to Urdu translation using Hierarchical Model given by (Chiang, et al., 2005) They also made a comparison of simple default phrase-based model with the hierarchical model and showed the performance of simple phrase-based is much better for such local language like Urdu then the hierarchical phrase-based approach to SMT.

(Singh et al., 2008) presented a Punjabi to Hindi Machine Translation System. The purposed system for Punjabi to Hindi translation has been implemented with various research techniques based on Direct MT architecture and language corpus. The output is evaluated in order to get the suitability of the system for the Punjabi Hindi language pair. A lot of work is being carried out using Neural Networks technology in the field of MT which is being a good approach nowadays. Neural Machine Translation is a newly proposed approach in MT. The main drawback using the approach is it requires relatively large amount of training corpus as compared to SMT. (Khalilov, et al., 2008) estimated a continuous space language model with a neural network in an Italian to English MT system. (Bahdanau,et al., 2014) presented a Neural Machine Translation by joint learning to align and translate. In our work we used Phrase-based SMT models and evaluated their performance on the morphologically rich Indian languages.

## 2. Evaluation

In this section, we discuss different datasets used in our experiments followed by discussion on training, tuning and testing of different model components and lastly results and related discussion.

### 2.1. Dataset

For this work, parallel corpora from diverse domains were collected for all the selected languages. For the bilingual corpus collection our first motive was to collect data from diverse domains to get better translation quality and a wide range vocabulary. For this purpose the corpus we selected to use in our work is EMILLE (Enabling Minority Language Engineering). EMILLE is a 63 million word corpus of Indic languages (Baker, et al., LREC 2002) which is distributed by the European Language Resources Association (ELRA). EMILLE contains data from six different categories: consumer, education, health, housing, legal and social documents. This data is based on the information leaflets provided by the UK government and various local authorities. We used 72 parallel files in total for each of our source language with each filename consisting of language code, text type (written or spoken), genre and subcategory, connected with hyphen character. The data is encoded in full 2-byte Unicode format and marked up in SGML format.

We used EMILLE corpus that is becoming a standard data repository for languages of this region. The parallel corpus consists of 200000 words of text in English and its accompanying translations in Hindi, Bengali, Punjabi, Gujarati and Urdu. Its bilingual resources consists of approximately 13000 sentences for all the available languages from which we were able to sentence-aligned and extract over 8000 sentence for all languages pairing with English using the sentence alignment algorithm given by (Moore., 2002). Some experiments with Multi-Indic parallel corpus (Post, et al. 2012) were also done.

Table 1: Training and Evaluation data for EMILLE

| Corpus | Total Sentences | Training Sentence | Tuning Sentence | Testing Sentence |
|---|---|---|---|---|
| **Bengali** | 8520 | 6816 | 852 | 852 |
| **Guajarati** | 8330 | 6664 | 833 | 833 |
| **Hindi** | 9510 | 7608 | 951 | 951 |
| **Punjabi** | 8465 | 6772 | 847 | 846 |
| **Urdu** | 8245 | 6596 | 825 | 824 |

In any SMT development project, development of parallel corpus is the most complicated task. In our work we have used EMILLE corpus which is quite noisy. Filtering and cleaning is requires before use. Cleaning of the corpus to extract aligned parallel sentences pair is the first step for development of any SMT system. Details about number of parallel sentences that were extracted for each pair are given in Table 1 and Table 2.

Table 2 : EMILLE Vocabulary Size for training and test **set**

| Source Language | Target Language (English) | | | | | |
|---|---|---|---|---|---|---|
| | Training Size (Tokens) | | Test Size (Tokens) | | Total Sentence Pairs (Tokens) including tuning sentence tokens. | |
| | Source | Target | Source | Target | Source | Target |
| **Bengali** | 98,952 | 90,523 | 11,073 | 10,123 | 124,745 | 113,923 |
| **Gujarati** | 89,995 | 86,594 | 10,328 | 9,785 | 112,676 | 107,695 |
| **Hindi** | 137,623 | 102,754 | 15,583 | 11,517 | 172,352 | 128,741 |
| **Punjabi** | 110,014 | 89,136 | 13,602 | 10,554 | 123,616 | 99690 |
| **Urdu** | 124,755 | 86,563 | 13,465 | 9,222 | 138,220 | 95785 |

A sufficiently large English language monolingual corpus is collected for this work. This monolingual corpus is used to build the language model that is used by the decoder to select the most affluent translation from several possible translation options. In this study we also tried to gather sufficiently large monolingual data from as many different available online resources as possible like *Europarl* (Koehn, 2005). The next step is to train the language model on the corpus that is suitable to the domain. To fulfil this need, data from diverse domains is collected. The main categories of the collected data are News, Religion, Health, Literature, Science, and Education. The WMT 08 News Commentary dataset is used as the main entity for monolingual data, the target side of the parallel corpora is also added to the monolingual data.

The monolingual corpora collected for this study have around 60 million tokens distributed in nearly 2 million sentences. These figures cumulatively present the number of tokens in all the domains whose data is used to build the language model. It includes monolingual data of the target languages of all parallel corpora collected for this study.

We also trained state-of-the-art phrase-based systems using the Multi-Indic parallel data that has been recently made available. It contains parallel data for six languages namely Bengali, Hindi, Malayalam, Tamil, Telugu and Urdu. The number of segments used for training, tuning and testing of different language pairs are shown in Table 3.

Table 3: Training and Evaluation Data for Indic Corpus

| Corpus | Training | Tuning | Testing |
| --- | --- | --- | --- |
| Bengali-English | 24000 | 775 | 1000 |
| Hindi-English | 39000 | 1000 | 1000 |
| Malayalam-English | 39000 | 1000 | 1000 |
| Tamil- English | 46000 | 1000 | 1000 |
| Telugu-English | 45000 | 1000 | 1000 |
| Urdu-English | 87000 | 980 | 883 |

## 2.2. Experimental Setup

For EMILLE corpus we performed k-fold cross validation method for sampling of the corpus for all language pairs. Here k=5 was selected by taking 4/5 of the total corpus as training and 1/5 as tuning and test set for experiment on all folds. Each fold comprises over 800 segments for tuning and same number of sentences for testing along with above 6500 segments for training for all source languages except Hindi. For Hindi we got above 9000 segments in total. Above 7000 selected for training and about 950 sentences for tuning and testing of Hindi to English translation system.

All these statistics can be seen clearly in Table 1. The first step in our work is sampling of data. Next, training, tuning and test sets are tokenized for all folds. Finally, all datasets are converted to lowercase. This process is repeated for all language pairs using scripts provided by Moses (Koehn et al. 2007) decoder. The lowercase training data is used for word alignment.

**Baseline Settings:** We trained a Moses system (Koehn et al., 2007) with the following features: a maximum sentence length of 80, GDFA symmetrization of GIZA++ alignments (Och & Ney, 2003), an interpolated Kneser-Ney smoothed 5-gram language model with SRILM (Stolcke, 2002) used at runtime, a 5-gram OSM (Durrani et al., 2013), msd-bidirectional-fe lexicalized reordering, sparse lexical and domain features (Hasler et al., 2012), a distortion limit of 6, 100-best translation options, MBR decoding (Kumar & Byrne, 2004), Cube Pruning (Huang & Chiang, 2007) with a stack-size of 1000 during tuning and 5000 during test, and the no-reordering-over punctuation heuristic. We tuned with the k-best batch MIRA algorithm (Cherry & Foster, 2012).

Language Model is built on the available monolingual English corpus. This language model is implemented as an n-gram model using the SRILM (Stolcke, 2002) toolkit. For all the experiments in all languages, the same language model is used for all folds of the source languages as translation is being performed from South Asian into English. For Multi-Indic experiments, we trained the language model using the monolingual WMT-13 data which is built from 148M English sentences.

## 2.3. Results

As the languages used in this work are sparse-resourced, we achieved relatively lower scores for BLEU (Papineni, 2002), we have achieved BLEU score with a mean of 0.12 and a Standard deviation of 0.06 on the given test sets using the 5-fold cross validation method. Table 4 presents the results of experiments for all language pairs. The results are composed of BLEU and NIST score evaluated over the test corpora and also the UNK (OOV Words) Count over that test corpus for all the selected language pairs. The subsequent subsections present evaluation results for all language pairs for both seen i.e. data taken from the training set and the unseen i.e. actual testing data.

Table 4 : Evaluation Results of developed SMT system for all language pairs

| Language Pair | BLEU | | NIST | | UNK Count | |
| --- | --- | --- | --- | --- | --- | --- |
| | Mean X | $\sigma$ | Mean X | $\sigma$ | Mean X | $\sigma$ |
| Bengali-English | 0.118 | 0.043 | 3.786 | 0.522 | 203 | 20 |
| Gujarati-English | 0.119 | 0.059 | 3.674 | 0.701 | 226 | 25 |
| Hindi-English | 0.115 | 0.068 | 3.779 | 0.804 | 224 | 30 |
| Punjabi-English | 0.150 | 0.09 | 4.185 | 1.158 | 197 | 36 |
| Urdu-English | 0.140 | 0.038 | 4.260 | 0.535 | 183 | 15 |

**Bangla-English:**

For Bengali-English language pair, we achieved decent BLEU scores with a mean of X= 0.118 and a Standard deviation σ = 0.043 on unseen data and X= 0.364 with a Standard deviation σ = 0.018 on seen data. For NIST we got, X= 3.786 and a Standard deviation σ = 0.522 on unseen data and X= 7.878 with a Standard deviation σ = 0.328 on seen data.

When counting the unknown words in translation of our SMT system we achieved X= 610 and a Standard deviation σ = 59 on unseen data and X= 130 with Standard deviation σ = 8 on seen data. An example of translation output from the trained system is given below. The example is composed of Source segment with its reference translation from test corpus. A segmented output of translation output is also given.

**Example:**

Source: ডিপার ○ টমেন ○ ট অফ দি এনভায়রণমেন ○ ট ট ○ রান ○ সপোর ○ ট এণ ○ ড দি রিজিওনস

Reference: department of the environment transport and the regions

Output: the department of |0-5| the environment |6-9| transport |10-16| and the regions |17-22|

The indexes in the output represent which source words produced this output for example, "the department of" was produced by a source phrase containing source words indexed between 0 and 5.

Table 5 : Bangla-English Phrase table for given example

| S.No | Input Phrase | Reference Phrase |
|---|---|---|
| 1 | ডিপার ○ টমেন ○ ট অফ | the department of |
| 2 | দি এনভায়রনমেন ○ ট | of the environment |
| 3 | ট ○ রান ○ সপোর | Transport |
| 4 | ট এণ ○ ড দি রিজিওনস | and the regions |

Table 5 presents input phrases along with corresponding reference phrases for the example mentioned above. A clear difference can be observed between the reference translation and the one achieved from the developed system. The translation output is segmented into different phrases and decoder fetches the translation from the developed phrase table. The reordering model also gave poor result for such small amount of data.

In output the first six words of source are translated to "The department of" then next three to "the environment" then next five to just a single output "transport" and so on. Here it can be noted that how sparseness affect the output, the phrase table contains only one single output word for five input words. Table 6 shows the actual BLEU, NIST score for all the folds along with the OOV words count.

Table 6 : Evaluation results for Bangla-English translation

| Folds | BLEU | | NIST | | UNK Count | |
|---|---|---|---|---|---|---|
| | Seen | Unseen | Seen | Unseen | Seen | Unseen |
| F1 | 0.403 | 0.075 | 8.284 | 3.157 | 129 | 630 |
| F2 | 0.342 | 0.082 | 7.590 | 3.36 | 120 | 670 |
| F3 | 0.347 | 0.098 | 7.517 | 3.826 | 141 | 617 |
| F4 | 0.363 | 0.153 | 7.899 | 4.249 | 122 | 621 |
| F5 | 0.375 | 0.182 | 8.101 | 4.338 | 138 | 512 |

**Gujarati-English:**
For this pair, again we got decent BLEU scores as compared to our small amount of training corpus with a mean of X= 0.119 and a Standard deviation σ = 0.059 on unseen data and X= 0.403 and a Standard deviation σ = 0.012 on seen data.

For NIST we got, X= 3.674 and a Standard deviation σ = 0.701 on unseen data and X= 8.136 and a Standard deviation σ = 0.153 on seen training corpus.

When counting the unknown words in translation of our SMT system we achieved X= 678 and a Standard deviation σ = 77 on unseen data and X= 117 and a Standard deviation σ = 16 on seen data.

An example of translation output from the trained system is given below. The example is composed of Source segment with its reference translation from test corpus. A segmented output of translation output is also given.

**Example:**

Source: અમુક બેનિફિટો માટે તમે નેશનલ ઇનશ ○ યોરન ○ શકોન ○ ટ ○ રીબ ○ યુશનો ભરેલાં હોવાં જ જોઇએ અથવા એવું માની લેવામાં આવશે કે તમે તે ભરેલાં છે .

Reference: for some benefits you must have paid or be treated as having paid no contributions.

Output: for some benefits you |0-4| ઇનશ |5-5| your |8-9| no |6-7| contributions |10-16| must |19-20| have paid |17-18| or |21-21| be |22-22| taken |24-24| to

Table 7 : Gujarati-English Phrase table for given example

| S.No | Input Phrase | Reference Phrase |
|---|---|---|
| 1 | અમુક બેનિફિટો માટે તમે નેશનલ | for some benefits you |
| 2 | ઇનશ | NULL |
| 3 | ચોરન | No |
| 4 | શકોન | Your |
| 5 | ટ ૦ રીબ ૦ યુશનો | Contributions |
| 6 | ભરેલાં હોવાં | have to have paid |
| 7 | જ જોઇએ | Must |
| 8 | અથવા | Or |
| 9 | માની લેવામાં આવશે | be deemed to have ceased |

Table 7 presents input phrases along with corresponding reference phrases for the example mentioned above. A clear difference can be observed between the reference translation and the one achieved from the developed system. The translation output is segmented into different phrases and decoder fetches the translation from the developed phrase table. The reordering model also gave poor result for such small amount of data.

In output the first four words of source are translated to "for some benefits you" then next word could not be translated by the decoder so it becomes an OOV in our translation output. From the phrase table it is seen that many source words translated to just single target output. This is also because of poor tokenization for regional languages as there is no standardized tokenizer available for these languages. Table 8 shows the actual BLEU, NIST score for all the folds along with the OOV words count.

Table 8 : Evaluation results for Gujarati-English translation

| Folds | BLEU | | NIST | | UNK Count | |
|---|---|---|---|---|---|---|
| | Seen | Unseen | Seen | Unseen | Seen | Unseen |
| F1 | 0.413 | 0.081 | 7.942 | 3.251 | 144 | 730 |
| F2 | 0.397 | 0.079 | 8.215 | 3.146 | 106 | 709 |
| F3 | 0.391 | 0.089 | 8.072 | 3.226 | 119 | 729 |
| F4 | 0.399 | 0.131 | 8.104 | 3.968 | 108 | 677 |
| F5 | 0.420 | 0.219 | 8.349 | 4.338 | 107 | 546 |

### Hindi-English:

The results of Hindi to English translation are given in Table 10. The corpora used for Hindi-English language pair was the most domain-relevant and the biggest in size. It resulted in significantly better translation as compare to other language pairs. Hence, it can be concluded that the size and relevance of parallel language corpus have a direct relationship with the quality of translation. For this pair, again we got decent BLEU scores with a mean of X= 0.115 and a Standard deviation σ = 0.068 on unseen data and X= 0.352 and a Standard deviation σ = 0.025 on seen data. For NIST we got, X= 3.779 and a Standard deviation σ = 0.804 on unseen data and X= 7.634 and a Standard deviation σ = 0.437 on seen data.

When counting the unknown words in translation of our SMT system we achieved X= 672 and a Standard deviation σ = 90 on unseen data and X= 150 and a Standard deviation σ = 10 on seen data. Translation output of our developed system is given below in example.

**Example:**

Source: उनसे समंपर ○ के लिए पते व टेलीफोन नंबर नीचे दिए हैं○:

Reference: contact addresses and telephone numbers are as follows:

Output: on |0-0| the |2-3| समंपर |1-1| for |4-5| addresses |6-6| and |7-7| telephone |8-8| helpline |9-9| below |10-10| दिए |11-11| : |12-12|

Table 9: Hindi-English Phrase table for given example

| S.No | Input Phrase | Reference Phrase |
|---|---|---|
| 1 | उनसे | On |
| 2 | क के | The |
| 3 | समंपर | NULL |
| 4 | के लिए | For |
| 5 | टेलीफोन | Telephone |
| 6 | पते व | Addresses |
| 7 | नीचे दिए | Of the following |
| 6 | हैं○: | : |
| 7 | नंबर | Helpline |

Table 7 presents input phrases along with corresponding reference phrases for the example mentioned above. A clear difference can be observed between the reference translation and the one achieved from the developed system. The translation output is segmented into different phrases and decoder fetches the translation from the developed phrase table. The reordering model also gave poor result for such small amount of data.

In output the first word of source is translated to "on" then next two words were translated as "the" then again NULL token so it becomes an OOV in our translation output. From the phrase table it is seen that many source words are translated to just single target output. This is also because of poor tokenization for regional languages as there is no standardized tokenizer available for these languages. Table 10 shows the actual BLEU, NIST score for all the folds along with the OOV words count.

Table 10 : Evaluation results for Hindi-English translation

| Folds | BLEU | | NIST | | UNK Count | |
|---|---|---|---|---|---|---|
| | Seen | Unseen | Seen | Unseen | Seen | Unseen |
| F1 | 0.365 | 0.065 | 7.765 | 3.531 | 134 | 701 |
| F2 | 0.381 | 0.074 | 8.036 | 3.076 | 155 | 735 |
| F3 | 0.366 | 0.068 | 7.396 | 3.483 | 160 | 754 |
| F4 | 0.323 | 0.151 | 7.677 | 5.114 | 154 | 637 |
| F5 | 0.328 | 0.221 | 7.910 | 5.721 | 149 | 533 |

**Punjabi-English:**

For this pair, again we got decent BLEU scores with a mean of X= 0.15 and a Standard deviation σ = 0.09 on unseen data and X= 0.385 and a Standard deviation σ = 0.053 on seen data.

For NIST we got, X= 4.185 and a Standard deviation σ = 1.158 on unseen data and X= 7.754 and a Standard deviation σ = 0.242 on seen data with relatively small amount of training parallel corpus.

When counting the unknown words in translation of our SMT system we achieved X= 591 and a Standard deviation σ = 110 on unseen data and X= 98 and a Standard deviation σ = 13 on seen data. The example given below composed of the input source with its reference from the parallel corpus and also the translation output from the developed system.

Example:

Source: ਪਹਿਲਾਂ ਇਹ ਪਤਾ ਕਰੋ ਕਿ ਤੁਹਾਨੂੰ ਕਿਹੜੇ ਬੈਨੀਫਿਟ ਮਿਲ ਸਕਦੇ ਹਨ ।

Reference: check first what benefit or benefits you may be able to get.
Output: check first |0-3| what |6-6| benefits |7-7| that |4-4| you |5-5| can get . |8-11|

Table 11: Punjabi-English Phrase table for given example

| S.No | Input Phrase | Reference Phrase |
|---|---|---|
| 1 | ਪਹਿਲਾਂ ਇਹ ਪਤਾ ਕਰੋ ਕਿ | Check first |
| 2 | ਜਿਹੜੇ | That |
| 3 | ਬੈਨੀਫਿਟ | Benefits |
| 4 | ਤੁਹਾਨੂੰ | You |
| 5 | ਮਿਲ ਸਕਦੇ ਹਨ। | can get . |

All the segments/phrases of source input are given in above phrase table of Table 11. We can find a number of differences between the reference and the translation output of the developed system. The translation output is segmented into different phrases and decoder fetches the translation from the developed phrase table. The reordering model also gave poor result for such small amount of data.

In output the first three words of source are translated to "check first" then all other words were translated to single words in output even the last phrase of over two to four words also translated to single word. From the phrase table it is seen that many source words translated to just single target output. This is also because of poor tokenization in pre-processing for regional languages as there is no standardized tokenizer available for these languages. In Table 12 we present actual BLEU, NIST score for all the folds along with the OOV word count.

Table 12: Evaluation results for Punjabi-English translation

| Folds | BLEU | | NIST | | UNK Count | |
|---|---|---|---|---|---|---|
| | Seen | Unseen | Seen | Unseen | Seen | Unseen |
| F1 | 0.409 | 0.099 | 7.913 | 3.094 | 121 | 677 |
| F2 | 0.397 | 0.071 | 8.065 | 3.348 | 94 | 703 |
| F3 | 0.346 | 0.095 | 7.871 | 3.172 | 93 | 615 |
| F4 | 0.369 | 0.205 | 7.177 | 4.445 | 89 | 522 |
| F5 | 0.408 | 0.283 | 7.147 | 4.839 | 94 | 440 |

**Urdu-English**

For this language pair, we got BLEU scores with a mean of X= 0.14 and a Standard deviation σ = 0.038 on unseen data and X= 0.371 and a Standard deviation σ = 0.027 on seen data. For NIST we got, X= 4.26 and a Standard deviation σ = 0.535 on unseen data and X= 7.54 and a Standard deviation σ = 0.53 on seen data with very small amount of training parallel corpus.

When counting the unknown words in translation of our SMT system we come up with X= 550 and a Standard deviation σ = 45 on unseen data and X= 117 and a Standard deviation σ = 12 on seen data. The example given below shows the different kind of problems that are occurred in getting translation output from the developed system.

Example:

Source: 20. ‎بہتری کی یہ باتیں ایک عمدہ ابتدا ہیں ۔
Reference: 20. These improvements are a good start.

Output: 20. |0-0| the |1-2| these |3-3| things to |4-4| start |7-7| a |5-5| quality |6-6| . |8-9| |||

Table 13: Urdu-English Phrase table for given example

| S.No | Input Phrase | Reference Phrase |
|---|---|---|
| 1 | 20. | Null |
| 2 | بہتری کی | the need for improvements |
| 3 | یہ | These |
| 4 | باتیں | Things to |
| 5 | ایک | A |
| 6 | عمدہ | Good quality |
| 7 | ابتدا | Start |
| 8 | ہیں ۔ | . |

In output the first word of source and target is same so decoder did nothing with it and its segment from phrase table will be NULL. The next word got totally different output in translation output as compared to the phrase table entry of Table 13. The two source words are translated to four word phrase in phrase table but in our translated output we got just a single output translation. This is because of the n-best translation phrase for a single phrase input. Next we can see the reordering again poorly managed by the baseline phrase based model.

All this discussion with given output example lead us to a bottom line conclusion that if we manage to get a good tokenizer and more corpora for all the selected regional languages, it will lead us to decent BLEU scores and fluent translations. Table 14 shows the actual BLEU, NIST score for all the folds along with the OOV word count.

Table 14: Evaluation results for Urdu-English translation

| Folds | BLEU | | NIST | | UNK Count | |
|---|---|---|---|---|---|---|
| | Seen | Unseen | Seen | Unseen | Seen | Unseen |
| F1 | 0.401 | 0.110 | 8.178 | 3.721 | 97 | 563 |
| F2 | 0.343 | 0.097 | 7.219 | 3.777 | 113 | 573 |
| F3 | 0.383 | 0.139 | 7.737 | 4.174 | 125 | 539 |
| F4 | 0.388 | 0.161 | 7.613 | 4.784 | 123 | 597 |
| F5 | 0.341 | 0.194 | 6.953 | 4.845 | 127 | 478 |

**Multi-Indic Corpus**

The results from running state-of-the-art baseline systems on multi-Indic corpus are shown in Table 15 below. For these experiments, we additionally transliterated OOV words following the unsupervised post-decoding transliteration method as described in (Durrani et al., 2014).

Table 15: Evaluation Multi-Indic Corpus

| Language | Tuning | Test |
|---|---|---|
| Bengali-English | 0.197 | 0.167 |
| Hindi-English | 0.193 | 0.16 |
| Malayalam -English | 0.111 | 0.09 |
| Tamil-English | 0.128 | 0.066 |
| Telugu-English | 0.142 | 0.110 |
| Urdu-English | 0.247 | 0.238 |

Increasing data can improve BLEU scores in all the language pairs we reported. However, the data available for Indian languages is still not enough to reliably estimate translation and reordering models. We can see clearly from Table 2 the vocabulary size is not good enough in numbers for training of the SMT system and it is creating data sparsity issue. More data is required to produce better translations. Translation quality can also be improved by studying the similarities between these languages. Data sparsity can be overcome by using methods of triangulation (Cohen and Lapata, 2007) & (Bertoldi et. al, 2008) and transliteration (Durrani et. al, 2010) which have been shown to be useful for closely related languages.

## 3. Future Work & Conclusion

The developed SMT system takes the Indian language sentences as input and it generates corresponding closest translation in English. The translation of over 800 sentences was evaluated using automatic evaluation metric i.e. BLEU evaluation. Average BLEU score of 10% to 20% was reported for all the languages. From these low BLEU scores it may be concluded that the quality of translation is directly dependent on the scope and quality of parallel language corpora.

In this work we introduced all the less explored language from India pairing with English. As all the eight Indian Languages used in this work exhibit rich morphology thus resulting in sparse estimates which causes poor translation quality, therefore our results are not as good as the ones reported for the European languages (Koehn et al., 2005) for which parallel and monolingual data is abundantly available.

In this work we employee Phrase-based model for training and used MERT for tuning our system. We carried out a set of experiments by choosing the training, tuning and test sets from parallel corpus using the fivefold cross validation method to make up the fact that we had only a small amount of parallel data. We found that each of our source Indian language got so much divergence when translating into English and that is why there is significant difference in obtained MT evaluation scores on seen corpus and on unseen test sets.

In future, we will study SMT by applying other different approaches to develop good language models and also the training model for all the South Asian languages whose more parallel corpus is available at the moment or may be available in nearer future. We also intend to perform a deep manual qualitative analysis on the MT output for all the language pairs we used for experimenting to compare our MT evaluation results for both the seen and unseen datasets as there are unknown words occurring in translation of seen test sets.

# References

Anwar MM, Anwar MZ, Md. Al-Amin Bhuiyan, (2009). Syntax Analysis and Machine Translation of Bangla Sentences, International Journal of Computer Science and Network Security,317–326.

Jawaid B and Daniel Zeman. (2011) Word-order issues in English-to-Urdu \Statistical Machine Translation. The Prague Bulletin of Mathematical Linguistics, (95):87– 106. ISSN 0032-6585.

Naskar SK, Sivaji Bandyopadhyay. (2006). A Phrasal EBMT System for Translating English to Bengali. Workshop on Language, Artificial Intelligence and Computer Science for Natural Language Processing applications

Singh G, (2008), "A Punjabi to Hindi Machine translation system". In proceeding of: COLING, 22nd International Conference on Computational Linguistics.

Islam, Z., Tiedemann, J. and Eisele, A. (2010). English to Bangla phrase-based machine translation. *In Proceedings of the 14th Annual conference of the European Association for Machine Translation*.

Cherry, C. and Foster, G. (2012). Batch tuning strategies for statistical machine translation. *In Proceedings of the 2012 Conference of the North American Chapter of the Association for Computational Linguistics:* Human Language Technologies, pages 427–436.

Chiang, D. (2005). A hierarchical phrase-based model for statistical machine translation, In *Proceedings of the 43rd Annual Meeting on Association for Computational Linguistics (ACL)*.

Dasgupta, S., Wasif, A. and Azam, S. (2004). An Optimal Way towards Machine Translation from English to Bengali, *Proceedings of the 7th International Conference on Computer and Information Technology (ICCIT)*.

Durrani, N., Sajjad, H., Fraser, A., and Schmid, H. (2010). Hindi-to-Urdu Machine Translation through Transliteration. In *Proceedings of the 48th Annual Meeting of the Association for Computational Linguistics*, pages 465–474.

Durrani, N., Fraser, A., Schmid H., Hoang, H. and Koehn, P. (2013). Can Markov Models Over Minimal Translation Units Help Phrase-Based SMT? In *Proceedings of the 51st Annual Meeting of the Association for Computational Linguistics*.

Durrani, N. and Hussain, S. (2010). Urdu word segmentation. In *Human Language Technologies: The 2010 Annual Conference of the North American Chapter of the Association for Computational Linguistics*, pages 528–536.

Durrani, N., Sajjad, H., Hoang, H. and Koehn., P. (2014). Integrating an Unsupervised Transliteration Model into Statistical Machine Translation. In *Proceedings of the 15th Conference of the European Chapter of the ACL*.

Hasler, E., Haddow B., and Koehn, P. (2012). Sparse lexicalised features and topic adaptation for smt. In *Proceedings of the seventh International Workshop on Spoken Language Translation*, pages 268–275.

Heafield, K. (2011). KenLM: Faster and Smaller Language Model Queries. *In Proceedings of the Sixth Workshop on Statistical Machine Translation*, pages 187–197.

Khan, N., Anwar, W., Bajwa, U. and Durrani, N. (2013). English to Urdu Hierarchical Phrase based SMT system. In *The fourth workshop n South and Southeast Asian NLP (WSSANLP), International Joint Conference on Natural Language Processing,* pages 72-76.

Post, M., Callison-Burch C. and Osborne.M (2012). Constructing Parallel Corpora for Six Indian Languages via Crowdsourcing. *In Proceedings of the Seventh Workshop on Statistical Machine Translation*, pages 401–409.

Koehn, P., Och., F. and Marcu, D. (2003), Statistical Phrase-BasedTranslation. In HLT-NAACL: *Conference combining Human LanguageTechnology conference series and the North American Chapter of theAssociation for Computational Linguistics* conference series, pages 48–54.

Koehn, P. (2010), A book on Statistical Machine Translation. Cambridge University Press.

Koehn P, Christof Monz, (2005), Shared task: statistical machine translation between European languages, Proceedings of the ACL Workshop on Building and Using Parallel Texts.

Koehn, P., Hoang, H., Birch, A., Callison-Burch, C., Federico, M., Bertoldi, N., Cowan, B., Shen, W., Moran, C., Zens, R., Dyer, C., Bojar, O., Constantin, A. and Herbst, E. (2007), Moses: Open source toolkit for statistical machine translation, In *Proceedings of the 45th Annual Meeting of the Association for Computational Linguistics Companion Volume Proceedings of the Demo and Poster Sessions.* Prague, Czech Republic: Association for Computational Linguistics, pp. 177–180.

Moore, R. (2002). Fast and accurate sentence alignment of bilingual corpora. In *Conference of the Association for MachineTranslation in the Americas.*

Och, F. (2003). A systematic comparison of various statistical alignment models. *Computational Linguistics, 29(1):19–51.*

Och F. J. and Ney H. (2000), Improved Statistical Alignment Models: In Proceedings of the 38th Annual Meeting of the Association for Computational Linguistics (ACL), pp. 440–447.

Roy, M., (2009). A Semi-supervised Approach to Bengali-English Phrase-Based Statistical Machine Translation, *Proceedings of the 22$^{nd}$ Canadian Conference on Artificial Intelligence.*

Shannon, Claude E. (1948) A mathematical theory of communication. Bell System Technical Journal, 27:379–423 and 623–656.

Sharma, N., Parteek Bhatia, Varinderpal Singh. (2010), English to Hindi Statistical Machine Translation, *International Journal in Computer Networks and Security*.

Stolcke, A,. (2002). SRILM - An extensible language modeling toolkit. *In Intl. Conf. Spoken Language Processing.*

Singh, D. et al (2012). "Modeling phrase based SMT for English to Hindi language", IRREST.

J. Eisner. (2003). Learning non-isomorphic tree mappings for machine translation. In Proceedings of the ACL Interactive Poster/Demonstration Sessions, 205–208.

Trever Cohn and Mirella Lapata. (2007). Machine Translation by Triangulation: Making Effective use of Multi-Parallel Corpora. *In the 45$^{th}$ Annual Meeting of the Association for Computational Linguistics,*

Bertoldi, N., Barbaiani, M., Federico, M., & Cattoni, R. (2008). Phrase-based statistical machine translation with pivot languages. In International Workshop on Spoken Language Translation Evaluation Campaign on Spoken Language Translation (IWSLT), pp. 143–149.

Nakov, P. and Tiedemann, J. (2012). Combining word-level and character-level models for machine translation between closely-related languages. *In Proceedings of the 50th Annual Meeting of the Association for Computational Linguistics* (Volume 2: Short Papers), pages 301–305.

Khalilov, Maxim et al. (2008) 'Neural Network Language Models for Translation with Limited Data', In Proceedings of 20th IEEE International Conference on Tools with Artificial, pp 445-451.

Kenji Yamada *and Kevin Knight*. (2001). *A syntax*-based statistical translation model. *In Proceedings* of the *39th* Annual *Meeting of* the ACL, pages 523–530.

D. Bahdanau, K. Cho, and Y. Bengio. (2014), Neural machine translation by jointly learning to align and translate. arXiv preprint arXiv:1409.0473.

T. Liu, W. Che, S. Li, Y. Hu, and H. Liu, (2005). "Semantic role labeling system using maximum entropy classifier," In Proceedings of CoNLL, pp. 189-192.

Bisazza, A. and Federico, M., (2010) Chunk-based verb reordering in VSO sentences for Arabic-English statistical machine translation, in Proceedings of the Joint Fifth Workshop on Statistical Machine Translation and Metrics MATR, WMT '10, pp. 235–243.

The Editors of Encyclopædia Britannica. "Bengali Language." *Encyclopedia Britannica Online*. Encyclopedia Britannica, n.d. Web. 15 June 2014.

The Editors of Encyclopædia Britannica. "Gujarati Language." *Encyclopedia Britannica Online*. Encyclopedia Britannica, n.d. Web. 18 June 2014.

The Editors of Encyclopædia Britannica. "Hindi Language." *Encyclopedia Britannica Online*. Encyclopedia Britannica, n.d. Web. 18 June 2014.

The Editors of Encyclopædia Britannica. "Punjabi Language." *Encyclopedia Britannica Online*. Encyclopedia Britannica, n.d. Web. 18 June 2014.

The Editors of Encyclopædia Britannica. "Telugu Language." *Encyclopedia Britannica Online*. Encyclopedi a Britannica, n.d. Web. 18 June 2014.

The Editors of Encyclopædia Britannica. " Malayalam Language." *Encyclopedia Britannica Online*. Encyclopedia Britannica, n.d. Web. 18 June 2014.

The Editors of Encyclopædia Britannica. "Urdu Language." *Encyclopedia Britannica Online*. Encyclopedia Britannica, n.d. Web. 18 June 2014.

The Editors of Encyclopædia Britannica. "Tamil Language." *Encyclopedia Britannica Online*. Encyclopedia Britannica, n.d. Web. 18 June 2014.

Baker P. EMILLE, (2002), A 70-Million Word Corpus of Indic Languages: Data Collection, Mark-up and Harmonization: In Proceedings of the 3rd Language Resources and Evaluation Conference, pp. 819-825, LREC'.

Koehn Philipp (2005). EuroParl: A Parallel Corpus for Statistical Machine Translation. Machine Translation Summit.

Huang, L. and Chiang, D. (2007). Forest rescoring: Faster decoding with integrated language models. In Proceedings of the 45th Annual Meeting of the Association of Computational Linguistics, pages 144–151.

Kumar, S. and Byrne, W. J. (2004). Minimum bayes-risk decoding for statistical machine translation. In HLT-NAACL, pages 169–176.

Stolcke.A,(2002). SRILM - an extensible language modeling toolkit. In Intl. Conf. Spoken Language Processing.

Papineni K. (2002). BLEU: A method for automatic evaluation of machine translation : In Proceedings of 40th Annual meeting of the Association for Computational Linguistics (ACL), pp. 311–318.